\documentclass[twoside,11pt]{article}
\usepackage{jmlr2e}

\usepackage[utf8]{inputenc}
\usepackage{bold-extra}

\usepackage{wrapfig}
\usepackage{enumitem}
\usepackage{todonotes}

\usepackage{amsmath}
\usepackage{xcolor}
\usepackage{tikz, pgfplots}
\pgfplotsset{compat=1.3}
\usetikzlibrary{arrows.meta}
\tikzstyle{arrow}=[-{Computer Modern Rightarrow[scale width=0.5]}]
\definecolor{myred}{RGB}{186,33,33}
\definecolor{myblue}{RGB}{28,0,207}

\usepackage[frozencache,cachedir=.]{minted}
\usepackage{soul}

\jmlrheading{1}{2000}{1-48}{4/00}{10/00}{00-000}{author list} 
\ShortHeadings{JsonGrinder and Mill}{Mandlík, Račinský, Lisý, Pevný} 

\setitemize{noitemsep,topsep=0pt,parsep=0pt,partopsep=0pt}
\setenumerate{noitemsep,topsep=0pt,parsep=0pt,partopsep=0pt}

\newcommand{\mill}{\href{https://github.com/CTUAvastLab/Mill.jl}{\scriptsize{\texttt{{Mill.jl\ }}}}}
\newcommand{\jsongrinder}{\href{https://github.com/CTUAvastLab/JsonGrinder.jl}{\scriptsize{\texttt{{JsonGrinder.jl\ }}}}}
\newcommand{\json}{\ensuremath{\mathcal{H}}}
\newcommand{\atoms}{\ensuremath{\mathcal{A}}}

\begin{document}

\title{Mill.jl and JsonGrinder.jl: automated differentiable feature extraction for learning from raw JSON data }
\author{\name Šimon Mandlík
\email simon.mandlik@avast.com
\AND
\name Matěj Račinský
\email matej.racinsky@avast.com
\AND
\name Viliam Lisý
\email viliam.lisy@avast.com
\AND
\name Tomáš Pevný
\email pevnak@protonmail.ch
}
\editor{}

\maketitle
\begin{abstract}%
  Learning from raw data input, thus limiting the need for manual feature engineering, is one of the key components of many successful applications of machine learning methods. 
  While machine learning problems are often formulated on data that naturally translate into a vector representation suitable for classifiers, there are data sources, for example in cybersecurity, that are naturally represented in diverse files with a unifying hierarchical structure, such as XML, JSON, and Protocol Buffers.
  Converting this data to vector (tensor) representation is generally done by manual feature engineering, which is laborious, lossy, and prone to human bias about the importance of particular features. 
  \mill and \jsongrinder is a tandem of libraries, 
  which fully automates the conversion. Starting with an arbitrary set of JSON samples, they create a differentiable machine learning model capable of infer from  further JSON samples in their raw form.
  In the spirit of the Julia language, the framework is split into two packages --- \mill implementing the hierarchical multiple instance learning paradigm, offering a theoretically justified approach for building machine learning models for this type of data, and \jsongrinder summarizing the structure in a set of JSON samples and reflecting it in a \mill model. 
\end{abstract}

\section{Motivation}

\begin{wrapfigure}{r}{0.51\textwidth}
\centering
\vspace{-0.8cm}
\begin{minted}[fontsize=\tiny, fontfamily=tt]{json}
{ "mac": "00:04:4b:a9:c1:f3",
  "ip": "192.168.1.122",
  "services": [{ "protocol": "udp", "port": 5353 },
               { "protocol": "tcp", "port": 6466 }],
  "upnp": [{ "model_name": "AirReceiver",
             "manufacturer": "SoftMedia Inc.",
             "model_description": "AirReceiver - Media Renderer",
             "services": ["urn:upnp-org:serviceId:AVTransport",
                          "urn:upnp-org:serviceId:RenderingControl"]},
           { "model_name": "SHIELD Android TV",
             "manufacturer": "NVIDIA",
             "services": []}],
  "mdns_services": ["_airplay._tcp.local.",
                    "_nv_shield_remote._tcp.local."]}
\end{minted}
\vspace{-0.8cm}
\caption{A part of JSON sample from the Device ID challenge~\citep{DeviceID}.}\label{fig:json}
\vspace{-0.8cm}
\end{wrapfigure}

The last decade has witnessed a departure from feature-engineering to end-to-end systems taking raw data as an input. 
It substantially reduced human effort and increased performance for example in image recognition~\citep{krizhevsky2017imagenet}, natural language processing~\citep{devlin2019bert}, or game playing tasks \citep{silver2017mastering}.  
There are plethora of algorithms (and libraries) for creating classifiers, regressors, and other models when raw input means tensor of a fixed dimension (images), sequences (text), or general graphs. In contrast, a lot of data used in the enterprise sector (e.g., exchanged by web services) are stored in hierarchically structured formats like JSONs, XML, Protocol Buffers~\citep{Varda2008}, or Message Pack~\citep{furuhashi_2010}. 
Let us refer to 
the data of this format Hierarchical Multiple Instance Learning (HMIL) data.
Its structure resembles a tree with leaves being strings, numbers, or other primitive types; and internal nodes forming either \textbf{arbitrarily long} lists of subtrees (e.g. \texttt{services} in Fig.~\ref{fig:json}) or \textbf{possibly incomplete} sets of key-value pairs (e.g. elements of \texttt{upnp} in Fig.~\ref{fig:json}).
HMIL data cannot be represented as fixed vectors without the laborious and lossy feature engineering, and it cannot be represented as plain sequences without losing the key information captured by its structure (e.g., leaf data types, irrelevance of ordering of key-value pairs).

\begin{wraptable}{r}{7.7cm}
    \scriptsize{
    \begin{tabular}{l|c|rrr}
                 & Sample & \multicolumn{3}{c}{Accuracy}\\
         Dataset & \multicolumn{1}{|c|}{Size} &  Default & Tuned & Comp. \\\hline
         \texttt{Device ID} & \textsf{0.1k-0.3M} & 0.932 & 0.971 & 0.967  \\    
         \texttt{EMBER}  & \textsf{3k-6M} & 0.948 & 0.996 & 0.974 \\
         \texttt{Mutagenesis}  & \textsf{4k-8k} & 0.886 & 0.932 & 0.912 \\
    \end{tabular}}
\end{wraptable}
To verify the generality of our framework, we have applied it to a range of \textbf{uncurated} datasets, modifying \emph{only} the path to the input data. 
In the \texttt{Device ID} challenge~\citep{DeviceID} hosted on \url{kaggle.com}, the samples originate from a network scanning tool. In \texttt{EMBER}~\citep{anderson2018ember}, the samples were produced by a binary file analyzer.
\texttt{Mutagenesis}~\citep{mutagenesis} is a smaller dataset with molecules trialed for mutagenicity on \textit{Salmonella typhimurium}.
The table shows that the default setting of our framework reaches a very good performance off-the-shelf, while further tuning allows reaching the performance of competing approaches (denoted Comp.) taken from~\cite{DeviceID}, ~\cite{aslan2020comprehensive}, and~\cite{guo2008learning} respectively.


\section{Background on Hierarchical Multiple Instance Learning}

The set of all possible HMIL data samples (denoted \json) is defined recursively. Any data type that can be conveniently represented as a fixed-size vector (i.e., integers, floats, strings\footnote{Represented as histograms of n-grams by default}, dates, categorical values) is an \textbf{atom}ic HMIL data sample from set $\atoms \subseteq \json$. More complex HMIL samples are created using two constructions: 
\textbf{sets} --  $\{x_1,x_2,\ldots, x_n\} \in \json$ for $x_i \in \json$;
and \textbf{dictionaries} -- $\{ (k_i,v_i) | i \in 1\dots k\} \in \json$ for $k_i \in \atoms, v_i \in \json$. 
Keys $k_i$ in the dictionaries are identifiers of properties with some semantic meaning (for example \texttt{mac}, \texttt{ip}, \texttt{services}) rather than carriers of information.

It is common to assume that samples in one dataset obey some fixed \textbf{schema},
which means that if data in a particular set/dictionary value are atoms, they are of the same type and if they are more complex samples, they follow the same sub-schema. Furthermore, the keys in each dictionary should come only from a limited set. These assumptions are not 
\begin{wrapfigure}{r}{0.5\textwidth}
\vspace{-0.5cm}
\begin{tikzpicture}[transform shape, scale=1.3, shift={(0, 0.5)}]

    \node (v1) at (0, -0.5) {
            \tikz{
                \draw[gray!40,line width=2.0mm] (0.0,0.1) -- (0.2,0.1);
                \draw[gray!40,line width=2.0mm] (0.0,0.3) -- (0.2,0.3);
                \draw[gray!80,line width=2.0mm] (0.0,0.5) -- (0.2,0.5);
                \draw[gray!80,line width=2.0mm] (0.0,0.7) -- (0.2,0.7);
                \draw[gray!80,line width=2.0mm] (0.0,0.9) -- (0.2,0.9);
                \draw[gray!30,line width=2.0mm] (0.0,1.1) -- (0.2,1.1);
                \draw[gray!30,line width=2.0mm] (0.0,1.3) -- (0.2,1.3);
                \draw[gray!70,line width=2.0mm] (0.0,1.5) -- (0.2,1.5);
                \draw[black!70, -] (-0.05, 0.4) -- (0.25, 0.4);
                \draw[black!70, -] (-0.05, 1) -- (0.25, 1);
                \draw[black!70, -] (-0.05, 1.4) -- (0.25, 1.4);
                \draw[] (0, 0) grid[step=0.2] (0.2, 1.6);
            }
        };

    \node (y) at (0, -1.85) {$y$}; 
    \draw[arrow] (0, -1.35) -- (0, -1.65);

    \begin{scope}[shift={(-0.1, 0.25)}]
        \node[anchor=center] at (-0.5, 0.3) {\tiny{\color{myred}{\texttt{"00:04:4b:a9:c1:f3"}}}};
        \draw[arrow] (-0.9, 0.15) -- (-0.05, -0.05) node [black, pos=0.4, below] {\scriptsize{$f_1$}};
    \end{scope}

    \begin{scope}[shift={(0.1, 0.25)}]
        \node[anchor=center] at (1.0, 0) {\tiny{\color{myred}{\texttt{"192.168.1.122"}}}};
        \draw[arrow] (0.9, -0.15) -- (0.05, -0.35) node [black, pos=0.4, below] {\scriptsize{$f_2$}};
    \end{scope}

    \begin{scope}[shift={(0.1, 0.05)}]
        \draw[arrow] (-0.7, -0.65) -- (-0.25, -0.65) node [black, midway, below] {\scriptsize{$g_1$}};
        \node[draw, inner sep=1, outer sep=0] at (-1.1, -0.65) {\scriptsize{$\frac{1}{n}\sum$}};
        \draw [decorate, decoration={brace}] (-1.6, -0.3) -- (-1.6, -1);

        \node (v2) at (-2.05, -0.45) {
                \tikz{
                    \draw[gray!50,line width=2.0mm] (0.0,0.1) -- (0.2,0.1);
                    \draw[gray!50,line width=2.0mm] (0.2,0.1) -- (0.4,0.1);
                    \draw[gray!80,line width=2.0mm] (0.4,0.1) -- (0.6,0.1);
                    \draw[gray!80,line width=2.0mm] (0.6,0.1) -- (0.8,0.1);
                    \draw[black!70, -] (0.4, -0.05) -- (0.4, 0.25);
                    \draw[] (0, 0) grid[step=0.2] (0.8, 0.2);
                }
            };


        \node (v3) at (-2.05, -0.85) {
                \tikz{
                    \draw[gray!50,line width=2.0mm] (0.0,0.1) -- (0.2,0.1);
                    \draw[gray!50,line width=2.0mm] (0.2,0.1) -- (0.4,0.1);
                    \draw[gray!80,line width=2.0mm] (0.4,0.1) -- (0.6,0.1);
                    \draw[gray!80,line width=2.0mm] (0.6,0.1) -- (0.8,0.1);
                    \draw[black!70, -] (0.4, -0.05) -- (0.4, 0.25);
                    \draw[] (0, 0) grid[step=0.2] (0.8, 0.2);
                }
            };

        \node at (-1.75, 0.2) {\tiny{\color{myblue}{\texttt{5353}}}};
        \node at (-2.35, 0.2) {\tiny{\color{myred}{\texttt{"udp"}}}};

        \draw[arrow] (-1.75, 0.1) -- (-1.85, -0.3);
        \node at (-1.6, -0.1) {\scriptsize{$f_3$}};
        \draw[arrow] (-2.35, 0.1) -- (-2.25, -0.3);
        \node at (-2.5, -0.1) {\scriptsize{$f_4$}};

        \node at (-1.75, -1.6) {\tiny{\color{myblue}{\texttt{6466}}}};
        \node at (-2.35, -1.6) {\tiny{\color{myred}{\texttt{"tcp"}}}};

        \draw[arrow] (-1.75, -1.5) -- (-1.85, -1);
        \node at (-1.55, -1.25) {\scriptsize{$f_3$}};
        \draw[arrow] (-2.35, -1.5) -- (-2.25, -1);
        \node at (-2.5, -1.25) {\scriptsize{$f_4$}};

    \end{scope}

    \begin{scope}[shift={(0.1, -0.15)}]
        \draw[arrow] (0.7, -0.75) -- (0.05, -0.95) node [black, pos=0.4, below] {\scriptsize{$g_2$}};

        \node[draw, inner sep=1, outer sep=0] at (1.1, -0.75) {\scriptsize{$\frac{1}{n}\sum$}};

        \draw [decorate, decoration={brace}] (1.6, -1.1) -- (1.6, -0.4);

        \node (v4) at (1.75, -0.55) {
                \tikz{
                    \draw[gray!50,line width=2.0mm] (0.0,0.1) -- (0.2,0.1);
                    \draw[black!70, -] (0.2, -0.05) -- (0.2, 0.25);
                    \draw[] (0, 0) grid[step=0.2] (0.2, 0.2);
                }
            };


        \node (v5) at (1.75, -0.95) {
                \tikz{
                    \draw[gray!50,line width=2.0mm] (0.0,0.1) -- (0.2,0.1);
                    \draw[black!70, -] (0.2, -0.05) -- (0.2, 0.25);
                    \draw[] (0, 0) grid[step=0.2] (0.2, 0.2);
                }
            };

        \node at (1.75, 0.1) {\tiny{\color{myred}{\texttt{"AirReceiver"}}}};
        \node at (1.75, -1.6) {\tiny{\color{myred}{\texttt{"SHIELD Android TV"}}}};

        \node[anchor=west] at (1.75, -0.55) {\scriptsize{\ldots}};
        \draw[arrow] (1.75, -0.05) -- (1.75, -0.4);
        \node at (1.95, -0.175) {\scriptsize{$f_6$}};

        \node[anchor=west] at (1.75, -0.95) {\scriptsize{\ldots}};
        \draw[arrow] (1.75, -1.45) -- (1.75, -1.1);
        \node at (1.95, -1.3) {\scriptsize{$f_6$}};
    \end{scope}

\end{tikzpicture}
\vspace{-0.6cm}
\caption{A sketch of a suitable model for processing the document in Figure~\ref{fig:json}.}\label{fig:model}
\vspace{-0.7cm}
\end{wrapfigure}
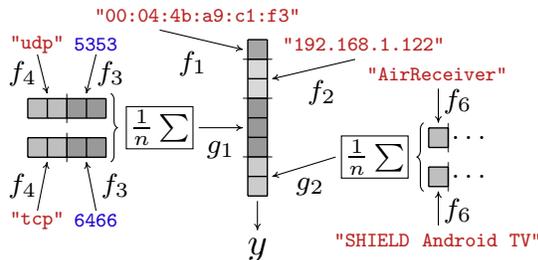
necessary for our framework to process the data, but they are necessary to achieve generalization to unseen samples. A schema can be derived automatically from a dataset.

The key idea of processing HMIL data is creating a hierarchy of embeddings, which gradually project atoms, sets, and dictionaries to fixed-sized vectors. The correctness of this approach is based on an extension of the universal approximation theorem~\citep{hornik1989multilayer} to HMIL data, proven in Theorem 5 by~\citet{pevny2019approximation}. The proof is recursive. For atoms, it holds from~\citep{hornik1989multilayer}; for sets $\{x_i | i\in 1\dots l\}$, any measurable function $f\colon 2^{\mathbb{R}}\rightarrow \mathbb{R}$ can be approximated to an arbitrary precision by a function $g(\frac{1}{l}\sum_{l=1}^{l} h(x_i)),$ where $g$ and $h$ are single hidden layer neural networks of an appropriate size. This construction was proposed by~\cite{pevny2017using,zaheer2017deep}. For dictionaries, concatenation of the embeddings of the values for a fixed ordering of keys creates a fixed-length vector, and hence the results by \citeauthor{hornik1989multilayer} are again applicable.

\section{Overview and Design}
The proposed framework implements the paradigm proposed by~\cite{pevny2017using,pevny2019approximation}.
It builds machine learning models for HMIL data in four steps:\footnote{The complete example is available at \newline \scriptsize{\href{https://github.com/CTUAvastLab/JsonGrinder.jl/blob/master/examples/mutagenesis.jl}{https://github.com/CTUAvastLab/JsonGrinder.jl/blob/master/examples/mutagenesis.jl}}. In total it has 85 lines of heavily commented code.} 
\begin{enumerate}
	\setlength\itemsep{0em}
	\item Deduce a \emph{schema} of a given HMIL dataset using the \texttt{JsonGrinder.schema} function.
	\item Create an \emph{extractor} for converting raw JSONs to internal structures of the \mill  \-library using \texttt{JsonGrinder.suggestextractor}. The implemented heuristics decide the correct representation for atoms (numbers, strings, categorical variables) and automatically correct bad practices in JSON formatting.
	\item Define a neural network model reflecting the schema using \texttt{Mill.reflectinmodel}.
	\item Train the model as any other model built using the \texttt{Flux.jl} library.
\end{enumerate}
The framework is written in the Julia language~\citep{Julia-2017} and split to two packages.

\vspace{2mm}
\noindent
Package \mill is built around the following invariant: Every node of HMIL data is represented (or wrapped) in a data node derived from the \texttt{AbstractDataNode} type. For every concrete realization of \texttt{AbstractDataNode}, there is a corresponding model (derived from \texttt{AbstractModel}), converting (embedding) the data node to a fixed-size vector.
This invariant ensures that data nodes can be arbitrarily nested (and so do corresponding models), because they know which data format to expect from child nodes. To increase space efficiency and allow for using efficient numerical libraries, all data for storing and processing a whole batch of samples are stored in continuous tensors.

\texttt{ArrayNode} is the data node representing atomic data, and it corresponds to \texttt{ArrayModel} that wraps a trainable function (usually a feed-forward neural networks (FNN)). \texttt{BagNode} represents sets and the corresponding \texttt{BagModel} implements various permutation invariant aggregation functions. A concatenation of coordinate-wise mean and maximum seems to be most effective in practice.
\texttt{ProductNode} represents dictionaries and the corresponding \texttt{ProductModel} contains a trainable function for each key. It applies them to the corresponding values, concatenates the outputs, and executes an additional trainable function on the concatenation.   
An important feature of \mill is that it can handle \emph{missing} data by explicitly storing them as \texttt{missing} values (a feature of Julia). In training and inference, the missing values are replaced by trainable imputations (unique for each node).

Since manually defining the hierarchy of model nodes for a given HMIL data can be a tedious and error-prone process, a function named \texttt{reflectinmodel} simplifies this task. For a given representative sample (without missing values), it creates a corresponding model, where the user prescribes only the constructor of the trainable functions given an input dimension, and the aggregation function to use in BagModels.


\vspace{2mm}
\noindent
\jsongrinder extends \mill to simplify the conversion of JSONs to its data structures.
Function \texttt{\textbf{schema}} calculates the data structure prevalent over a set of JSONs and basic statistics about its elements. For example, how often is a particular element present, distribution of lengths of lists at a specific position, and the distribution of leaf values. This information is useful to understand the data and it can even be visualized in HTML.
\textbf{Extractors} perform the conversion of JSONs to \mill nodes. JSON lists (see \texttt{"services"} in Fig.~\ref{fig:json}) are converted to \texttt{BagNode}s\footnote{This ignores the information contained in the list's ordering, but results in much more computationally efficient training. Support for sequences can be achieved by recurrent neural networks or transformers as shown in one of the examples in \mill, but this never achieved performance gains worth the computational cost in our experiments.} and JSON dictionaries (elements of \texttt{"upnp"} in Fig.~\ref{fig:json}) are mapped to \texttt{ProductNode}s. There are many ways to represent JSON leaves. \jsongrinder represents numbers directly, diverse collection of strings as n-gram histograms, and small collections of unique values as one-hot encoded categorical variables. The extractor can be created automatically from a schema using function \texttt{suggestextractor}, which uses heuristics to decide how to represent individual leaves. If the default extractors are not satisfactory, they can be easily replaced by custom implementations for particular schema nodes.

\subsection*{Integration with the ecosystem}
In the spirit of Julia package ecosystem, \mill provides just the required functionality while integrating well with other packages. This means that it uses \texttt{Flux.jl} for implementation of neural networks, any automatic differentiation engine like \texttt{Zygote.jl}, and it implements the interface of \texttt{LearnBase.jl} allowing to use other packages for dividing samples in minibatches, etc. \mill and \jsongrinder libraries are registered in the Julia default repository and can be added by the \texttt{Pkg.add} command.


\section{Conclusion}
\mill and \jsongrinder facilitate automated creation of models from HMIL data, which albeit ubiquitous in industry are rarely considered in machine learning literature. Both libraries are flexible, extensible, and well-integrated into the Julia ecosystem, allowing them to benefit from its improvement. Authors have used them on practical applications on big problems with $10^8$ samples of size up to 1GB each, frequently achieving better performance than with hand-designed features. Since we are not aware of any other software package that would allow the processing of JSON data samples without feature engineering, we consider the presented libraries to be an interesting contribution to the Auto-ML domain.

\end{document}